\documentclass{WileyMSP-template}
\usepackage{cite}
\usepackage{amsmath,amssymb,amsfonts}
\usepackage{algorithm}
\usepackage{algorithmic}
\usepackage{graphicx}
\usepackage{textcomp}
\begin{document}

\pagestyle{fancy}
\rhead{\includegraphics[width=2.5cm]{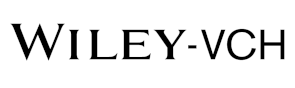}}

\title{Continual Learning for Multimodal Data Fusion of a Soft Gripper}

\maketitle

% Author: Please give full first and last names for authors and include * after the name of all corresponding authors

\author{Nilay Kushawaha*}
\author{Egidio Falotico}

% % Dedication

% \dedication{Optional dedication here. If no dedication is required, please leave blank}

% Affiliations: Please provide adacemic titles (Prof. or Dr.) for all authors where applicable, and include an institutional email address for all corresponding authors
\begin{affiliations}
Nilay Kushawaha\\
Address[1]: The BioRobotics Institute, Scuola Superiore Sant’Anna, Pontedera 56025, Italy\\
Address[2]: Department of Excellence in Robotics and AI, Scuola Superiore Sant’Anna, Pisa 56127, Italy
Email Address: nilay.kushawaha@santannapisa.it

Egidio Falotico\\
Address[1]: The BioRobotics Institute, Scuola Superiore Sant’Anna, Pontedera 56025, Italy\\
Address[2]: Department of Excellence in Robotics and AI, Scuola Superiore Sant’Anna, Pisa 56127, Italy

\end{affiliations}

% Keywords: Please provide a minimum of three and a maximum of seven keywords, separated by commas


% Abstract should be written in the present tense and impersonal style (i.e., avoid we), and be at most 200 words long
\begin{abstract}
Continual learning (CL) refers to the ability of an algorithm to continuously and incrementally acquire new knowledge from its environment while preserving previously learned information. Traditional models trained on a single data modality often fail to generalize when exposed to different modalities. A simple strategy to address this limitation is to fuse multiple modalities by concatenating their features and training the model on the combined representation. However, this generally requires re-training the model from scratch whenever a new domain is introduced. In this paper, we propose a continual learning algorithm that incrementally learns from multiple data modalities by combining class-incremental and domain-incremental learning settings. We validate our algorithm on a custom multimodal dataset composed of tactile signals collected from a soft sensorized pneumatic gripper and visual data consisting of non-stationary object images captured from video sequences. Additionally, we evaluate our method on a subset of the publicly available VGGSound dataset, which integrates both visual and audio signals corresponding to various real-world activities. To further demonstrate the robustness and real-time applicability of our approach, we conduct an object classification experiment using a soft sensorized gripper and an external camera setup, all synchronized through the Robot Operating System (ROS) framework.
\end{abstract}

\section{Introduction}
\label{sec:introduction}
In recent years, machine learning models have matched or even exceeded human-level performance on various tasks, such as image classification \cite{ahmad2021hyperspectral}, object recognition \cite{zou2023object}, natural language processing (NLP) \cite{otter2020survey}, and simulated game play \cite{silver2017mastering}. Despite their success in static settings where data distributions remain stable, these models struggle to adapt to evolving environments and often require re-training the network from scratch when new data becomes available. In contrast, humans demonstrate remarkable adaptability, continually learning, updating, and applying knowledge over time. Our goal is to develop AI systems that can replicate this level of continual adaptability.

Continual learning (CL) \cite{wang2024comprehensive}, also referred to as lifelong or incremental learning is a subfield of machine learning that focuses on training models progressively on a stream of data, allowing it to accumulate and retain knowledge over time. However, a major challenge for such models is their susceptibility to catastrophic forgetting of the previously acquired skills when learning new tasks/experiences or encountering shifts in data distribution. This often results in significant performance degradation, as new information overwrites existing knowledge. A straightforward but impractical solution is to store all past data, shuffle it, and re-train the model offline from scratch. While this can mitigate forgetting, it is computationally expensive, inefficient, and unsustainable for large-scale models. The goal of continual learning is to develop more scalable and efficient approaches that can maintain performance over time.

A model trained in one domain often performs poorly when deployed in a completely different domain or when applied to a different data modality than it was originally designed for. A common approach to address this issue is to either train separate models for each modality or to jointly train a single model offline on all modalities. However, these strategies are impractical for agents with limited memory and computational resources, especially those that learn new domains through active exploration and interaction with their environment. Another significant challenge is the scarcity of labeled data for supervised incremental learning. In contrast, unlabeled data is typically more abundantly available in real-world applications. To address this imbalance, semi-supervised learning (SSL) \cite{van2020survey} has emerged as a promising solution. In this setting, an agent leverages a small set of labeled data to extract useful knowledge, which can then be applied to learn new instances of the same class or to enhance it's understanding of previously encountered classes using the unlabeled data.

In this paper, we consider a realistic CL problem where a single CL model learns different modalities of data incrementally by leveraging both class-incremental and domain-incremental learning scenarios in an artificial environment where labeled data is scarce yet non-iid (independent and identical distribution) unlabeled data from the environment is plentiful. We also experiment with two different SSL conditions: \textit{Radom images}, where for each object of the same parent class we randomly sample some images from the dataset that are not included in the training data; \textit{Unique objects}, where we sample the images of new unseen objects that belong to the same parent class. More details about the SSL conditions is provided in section \ref{eval_metric}.

For our multimodal CL experiment, we employ two distinct types of dataset. The first setup involves tactile input from sensors embedded within a soft gripper, alongside visual data captured by an independent camera system. For the second experiment, we use the VGGSound dataset \cite{chen2020vggsound}, which offers paired image and audio modalities extracted from video clips. Our CL model is designed to incrementally learn class representations across these different domains. To address the challenges associated with multimodal CL, we build upon the Feature Covariance-Aware Metric (FeCAM) algorithm introduced by \cite{goswami2024fecam}. We propose an enhanced version, referred to as exFeCAM (extended FeCAM), which incorporates an online, task-wise semi-supervised learning strategy. Furthermore, our approach introduces an intra-layer feature representation mechanism that enables the learning of more generalized feature maps for each class. To the best of our knowledge, this is the first work to apply continual learning in a multimodal setting, enabling an agent to learn from different modalities in a domain-incremental manner. The key contributions of this work are as follows:

\begin{enumerate}
    \item We present an online non-exemplar CL algorithm with SSL capabilities. In addition, we incorporate intra-layer feature representation technique to obtain generalized feature maps. 
    \item We present a new multimodal non-iid dataset for real-world CL applications.
    \item We demonstrate the robustness of learning different modalities as a new incremental domain rather than just combining the data from different modalities into a single fused feature vector.
    \item We evaluate our algorithm on the custom multimodal dataset as well as the VGGSound dataset \cite{chen2020vggsound} and perform an ablation study to verify the effectiveness of the various parts of the algorithm. 
    \item We also perform a real-time experiment for object classification using a soft pneumatic gripper equipped with sensors and a separate camera setup, all synchronized using the ROS framework \cite{koubaa2017robot}. More details about the experiment is provided in section \ref{ros_application}.
\end{enumerate}
The paper is structured as follows: In Section 2, an overview of the related works in CL, SSL, and multimodality is presented, with a focus on the recent advancements in the literature. Section 3 provides a detailed description of the experimental setup that has been adopted discussing the different CL strategies, data accumulation steps, and evaluation metrics. Following this, the proposed methodology is presented in Section 4, which includes the model architecture and the framework used in this study. The evaluation of the proposed methodology on the custom multimodality dataset, as well as the VGGSound dataset is provided in section 5. Moreover, in section 6 the paper showcases the real-time application of the proposed algorithm. Finally, a summary noting the advantages and disadvantages of the proposed method along with the scope for improvements and possible developments in the future are stated in the last section.

\section{Related Work}
\subsection{Continual Learning Strategies}
Continual learning strategies can be broadly categorized into four major types: architectural, regularization, rehearsal, and hybrid strategies, each with it's own pros and cons. Architectural strategy either employs a dynamic modular network where new layers gets added upon encounter of new tasks \cite{rusu2016progressive}, \cite{fernando2017pathnet} or performs parameter isolation to allocate a dedicated separate sub-space for each task within the same network \cite{mallya2018piggyback}, \cite{serra2018overcoming}. However, a major drawback of this strategy is the increase in the size of the architecture as new tasks are learned or the potential saturation of the architecture if a fixed model is used.

Regularization strategies can be grouped into two: weight regularization and function regularization. Weight regularization constrains the update of the network weights by adding explicit regularization terms to the loss function, penalizing changes in each parameter based on it's importance \cite{kirkpatrick2017overcoming}, \cite{chaudhry2018riemannian}. Function regularization, on the other hand, employs distillation methods to maintain the knowledge between the various layers of the network. This typically involves treating the previously trained model as a teacher and the current model as a student, applying knowledge distillation (KD) techniques to align their outputs \cite{hinton2015distilling}, \cite{li2017learning}. However, regularizing the model can make it more difficult for the network to acquire new knowledge.

Rehearsal \cite{bang2021rainbow}, \cite{kumari2022retrospective}, \& pseudo-rehearsal, \cite{shin2017continual}, \cite{wu2018memory} approaches offer the most simple yet effective approach by storing a subset of the previously learned data in a memory buffer or by training a generative model to account for the previous information and replaying it during training on new tasks. However, storing the images in a replay buffer for rehearsal can raise privacy concerns. While using a generative model somewhat addresses the privacy issues, it may result in increased training time and computational demands.

Hybrid strategies \cite{van2020brain}, \cite{liu2020generative} are a combination of any of the other three strategies to compensate for catastrophic forgetting. SynapNet \cite{kushawaha2024synapnet} uses a triple model architecture along with a VAE-based pseudo-episodic memory for rehearsal and a sleep phase for memory re-organization. DualNet \cite{pham2021dualnet} uses a fast learner for supervised learning and a slow learner for unsupervised learning of task-agnostic general representation using semi-supervised learning. 

Non-exemplar CL strategies \cite{goswami2024fecam}, \cite{smith2021always} use the idea of replay-free sequential training, instead of using a modular network or regularizing the optimization function. These approaches store the prototypes of the past and new classes to compensate for catastrophic forgetting of the previously acquired knowledge. A pre-trained feature extractor provides the feature maps which are then used to calculate the mean and the covariance matrix for the respective classes.
\subsection{Semi-Supervised Learning}
SSL describes a class of algorithms that aim to learn from a combination of labeled and unlabeled data, which are typically assumed to be drawn from the same underlying distributions \cite{yan2023dml}. A central challenge in SSL lies in effectively leveraging both labeled and unlabeled data while mitigating the risks of overfitting, reducing the impact of noisy data, and ensuring meaningful improvements in model performance. Early advances in SSL using deep neural networks were driven by generative models such as autoencoders \cite{rasmus2015semi}, variational autoencoders \cite{kingma2014semi}, and generative adversarial networks \cite{odena2016semi}, which played a crucial role in extracting robust representations from partially labeled datasets.

Some SSL approaches \cite{oliver2018realistic}, \cite{berthelot2019mixmatch} focus on balancing two complementary loss terms: a supervised loss ($l_s$) applied to labeled samples and an unsupervised consistency regularization loss ($l_{ul}$) computed on the unlabeled data. These approaches are designed to ensure that the model learns not only from explicit supervision but also from consistency in predictions across augmented versions of unlabeled inputs. Two additional important approaches which have shown success in the context of deep learning are: self-training and conditional entropy minimization. Self-training, also known as pseudo-labeling \cite{lee2013pseudo}, involves assigning labels to unlabeled data based on the predictions of a model initially trained on labeled examples. In contrast, conditional entropy minimization \cite{grandvalet2004semi} encourages the model to make confident predictions on unlabeled samples by minimizing the entropy of their predicted class distributions. Although this does not ensure correctness due to the absence of true labels, it promotes decisiveness and reduces model uncertainty. In this work, we adopt the strategy of using cosine similarity between the feature representations of labeled and unlabeled data as a criterion for pseudo-labeling of the unlabeled samples. We refer to \cite{van2020survey} for a comprehensive review on the topic.

\subsection{Multimodality}
Our experience of the world is inherently multimodal we perceive objects through vision, hear sounds, feel textures, smell odors, and taste flavors. To develop an AI agent with similar capabilities, several approaches have been explored \cite{donato2024multi}, \cite{babadian2023fusion}. The simplest method to compensate for multiple domains is using the concept of joint representation where the unimodal signals are combined in the same representation space by simple concatenation \cite{baltruvsaitis2018multimodal}. \cite{ngiam2011multimodal} employed stacked denoising autoencoders to represent each modality individually and subsequently fusing them into a unified multimodal representation using an additional autoencoder layer.

Similarly, \cite{silberer2014learning} suggested a multimodal autoencoder for semantic concept grounding. They incorporated a reconstruction loss for training the representation and added a term in the loss function to predict object labels using the representation.\cite{kim2013deep} used a deep belief network (DBN) for each modality and then combined them into joint representations for audio-visual emotion recognition. A similar architecture was adopted by \cite{huang2013audio} for audio-visual speech recognition (AVSR), \cite{wu2014multimodal} applied the same strategy for integrating audio and skeletal joint representations in gesture recognition. \cite{sarfraz2024beyond} recently utilized a multimodal CL dataset to learn complementary information from two different modalities and performed a weighted fusion of the feature maps obtained from the respective feature extractors. On the other hand, \cite{xu2023towards} developed an egocentric multimodal dataset using smart glasses equipped with sensors and utilized the temporal binding network (TBN) \cite{kazakos2019epic} to fuse information from multiple sensors into a unified feature vector for feature extraction and training.

However, a key limitation of these methods is their inability to incrementally adapt to new domains. Specifically, when training on multiple domains (such as sensor signals, visual images), the addition of a new modality often requires re-training the entire model from scratch. In contrast, our proposed exFeCAM algorithm addresses this challenge by treating each modality as a distinct domain, allowing for the incremental integration of new modalities without the need to restart training. Furthermore, our approach eliminates the need to store data in memory for rehearsal, effectively addressing privacy concerns and the storage constraints commonly associated with exemplar-based CL methods.
\begin{figure*}
	\centering
	\includegraphics[width=\textwidth]{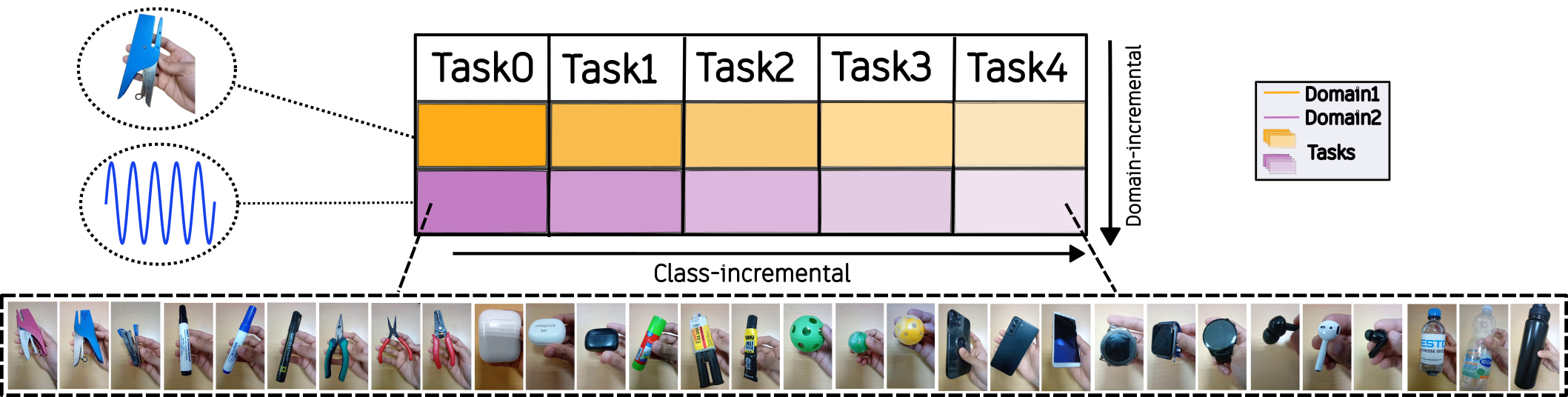}
	\caption{Multimodal data comprising of two different domains, domain 1 contains the images of various objects in different orientations, lightning conditions, and different levels of occlusion by the operator and domain 2 contains the tactile sensor information for the corresponding objects. Moving from left to right the model trains in a class-incremental manner, while going from top-bottom it trains in a domain-incremental manner.}
	\label{fig:multimodalData}
\end{figure*}

\section{Experimental Setup}
In this section, we describe the steps involved in the creation of the multimodal continual learning dataset as well as the standard evaluation criteria for the class-incremental as well as domain-incremental learning scenarios. 
\subsection{Continual Learning Scenario}
According to the division of incremental batches and the availability of task identities, continual learning can be broadly classified into three primary types or scenarios \cite{van2022three}:
\begin{itemize}
    \item Task-Incremental Learning (TIL): It requires an algorithm to learn a set of distinct tasks progressively, and it always knows which task it is working on.
    \item Class-Incremental Learning (CIL): The algorithm must learn to differentiate between a growing number of classes incrementally. 
    \item Domain-Incremental Learning (DIL): The algorithm needs to learn the same task, but in different a domain or environment.
\end{itemize}
In addition to these major scenarios, there are also other scenarios present in the literature like task-free continual learning (TFCL), online continual learning (OCL), blurred boundary continual learning (BBCL), continual pre-training (CPT), we refer to \cite{wang2024comprehensive} for a thorough study. In our experiment, we designed a hybrid scenario that integrates both class-incremental and domain-incremental learning paradigms.
\begin{figure}
	\centering
	\includegraphics[width=0.5\textwidth]{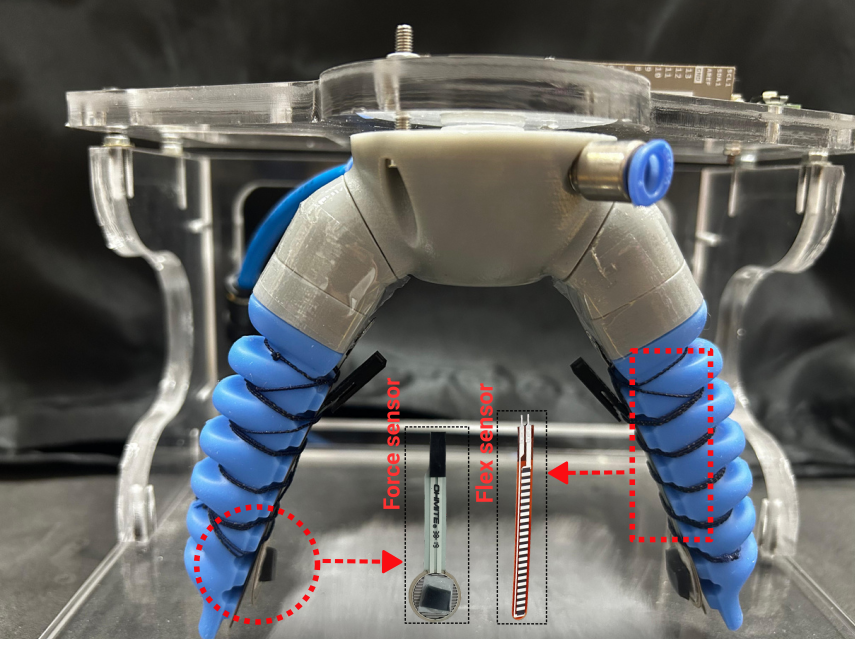}
	\caption{Commercial two-finger pneumatic gripper equipped with two force sensors and two flex sensors affixed together using stretchable nylon thread. The flex sensor measures the bending of the finger and the force sensor measures the magnitude of the force applied to the fingertip.}
	\label{fig:gripper}
\end{figure}

\subsection{Multimodal Dataset}
The multimodal dataset comprises two components: sensor signal from an actuated finger and non-stationary images of objects captured by a vision camera. The two different modalities are structured into a CL setup that incorporates both class-incremental as well as domain-incremental scenarios as illustrated in Figure \ref{fig:multimodalData}. In this setup, progression from left to right represents class-incremental training, while progression from top to bottom represents domain-incremental training. Both domains encompass five tasks, with each task involving the learning of two new classes. The first domain focuses on visual data for various objects encountered in each experience and follows a class-incremental learning approach. The second domain deals with sensor signal data of the same objects and employs both domain-incremental and class-incremental learning methods. Further details on data collection for the different modalities are provided in the subsequent sections.

In addition to our custom dataset, we also evaluate the effectiveness of our proposed method on a subset of the publicly available VGGSound dataset. We used this dataset due to it's multimodal capabilities and it's suitability for SSL settings. Specifically, we select five high-level parent classes: home, music, people, sports, and vehicles, each containing two specific subclasses. Every subclass includes 15 unique youtube videos for training, with each video lasting 240 seconds. Additionally, five separate videos per subclass are sampled in a similar fashion for testing. From these videos, we extract both the visual frames and the corresponding audio signals. To ensure accurate alignment between visual and audio modalities, we adopt the synchronization procedure outlined in \cite{sarfraz2024beyond}. Finally, the extracted visual frames are resized to 32x32 pixels to reduce computational complexity and accelerate the training process.

\subsubsection{Sensor Signal}
To collect sensor signals corresponding to different objects, we employ a soft pneumatic gripper \cite{hao2016universal} equipped with four sensors, which are secured in place using stretchable nylon thread. For capturing the bending motion of the fingers, we utilize commercially available flex sensors \cite{saggio2015resistive}, \cite{mishra2021recent}, while fingertip force measurements are obtained using force-sensitive resistors (FSRs) \cite{hollinger2006evaluation}, as illustrated in Figure \ref{fig:gripper}. The outputs generated by both the flex and force sensors are routed to an Arduino Due board, which interfaces with a computer for subsequent signal processing and data analysis. We collect sensor readings across 10 different object classes, with each object contributing 50 individual data points that reflect variations in object orientation. During the data acquisition phase, the gripper holds and releases each object multiple times at different contact positions, with each cycle spanning approximately 39 seconds. Every data point is represented as a 600-dimensional vector that encapsulates the full set of sensor signals recorded within that duration. The complete data acquisition process for each object requires approximately 37 minutes and is carried out on a workstation equipped with an Intel Core i7 CPU @2.80GHz and a (GTX 1070 GPU).

\subsubsection{Image Data}
The image data consists of the same classes but repeated with three different objects. Each object is repeated 3 times to create a more diversified dataset.  Classification can be performed at object level (30 classes) or at category level (10 classes). For each trial, the images are extracted from a 15-second video sequence at a rate of 10 frames per second (FPS). Objects are hand-held by the operator and the camera point-of-view is that of the operator's eyes. The operator extends his arm and smoothly rotates the object in front of the camera. The grabbing hand (left or right) changes throughout the sessions and relevant object occlusions are often produced by the hand itself. This creates a complex non-stationary dataset with variable lighting conditions and arbitrary amounts of occlusion produced by the operator's hand. The final dataset consists of 13,500 RGB images each of size 128x128 extracted from the video frame. The images are further resized to 32x32 dimension to speed up the training and evaluation process.  

\subsection{Evaluation Metrics and Benchmark Algorithms}\label{eval_metric}
The performance of the CL algorithm depends on it's ability to adapt, retain, and generalize the knowledge over time. We evaluate the incremental accuracy (denoted as $a_t$) of the model after every task as well as the average accuracy over all the tasks for each domain denoted as $A_T$. The average accuracy over all the domains is referred to as $A_D$ in the paper. We define the expressions as follows:\\
\begin{equation}
    A_T = \frac{1}{T}\sum_{t=1}^{t=T}a_t \hspace{1cm}
    A_D = \frac{1}{D}\sum_{T=1}^{T=D}A_T
\end{equation}
where $T$ and $D$ refer to the total number of incremental tasks in each domain and the total number of domains.\\
Additionally, to evaluate the SSL capability of our algorithm we perform two separate experiments as proposed in the paper \cite{smith2021memory}, we use the following terminology to describe them: First, \textit{Random Images}, where we have 30\% of unlabeled data and 70\% labeled data for each class. The unlabeled data is obtained by randomly sampling instances from the training dataset (ensuring that these images are removed from the labeled training set), removing their class labels, shuffling them, and storing them separately. Second, \textit{Unseen Images}, where we sample the images of an unseen object of the same parent class from the training dataset and repeat the above steps. A point to note is that in the first case, the model might have already seen all the objects for each class however, for the second case, the model has never seen the objects present in the unsupervised dataset which makes it a more challenging situation.

To assess the effectiveness of the proposed exFeCAM algorithm, we compare it against five benchmark methods: naive-concat, naive-add, vanilla FeCAM, FeTrIL, and a joint training approach. The naive-concat method employs a non-CL strategy by concatenating feature maps from both modalities for each class, followed by incremental training on newly introduced objects. In contrast, naive-add applies the same incremental learning procedure but replaces concatenation with a weighted summation of the feature maps. Both naive-concat and naive-add utilize an identical 3-layer feedforward neural network architecture, but without any CL capability. Vanilla FeCAM corresponds to the original FeCAM architecture introduced by \cite{goswami2024fecam}. FeTrIL \cite{petit2023fetril} also follows a replay free approach and stores the centroid representations of the past classes to produce their pseudo features by using geometric translation when training on new classes/domains. Lastly, the joint training approach involves offline training on the entire dataset, establishing an upper bound on accuracy.
\begin{figure*}
	\centering
	\includegraphics[width=\textwidth]{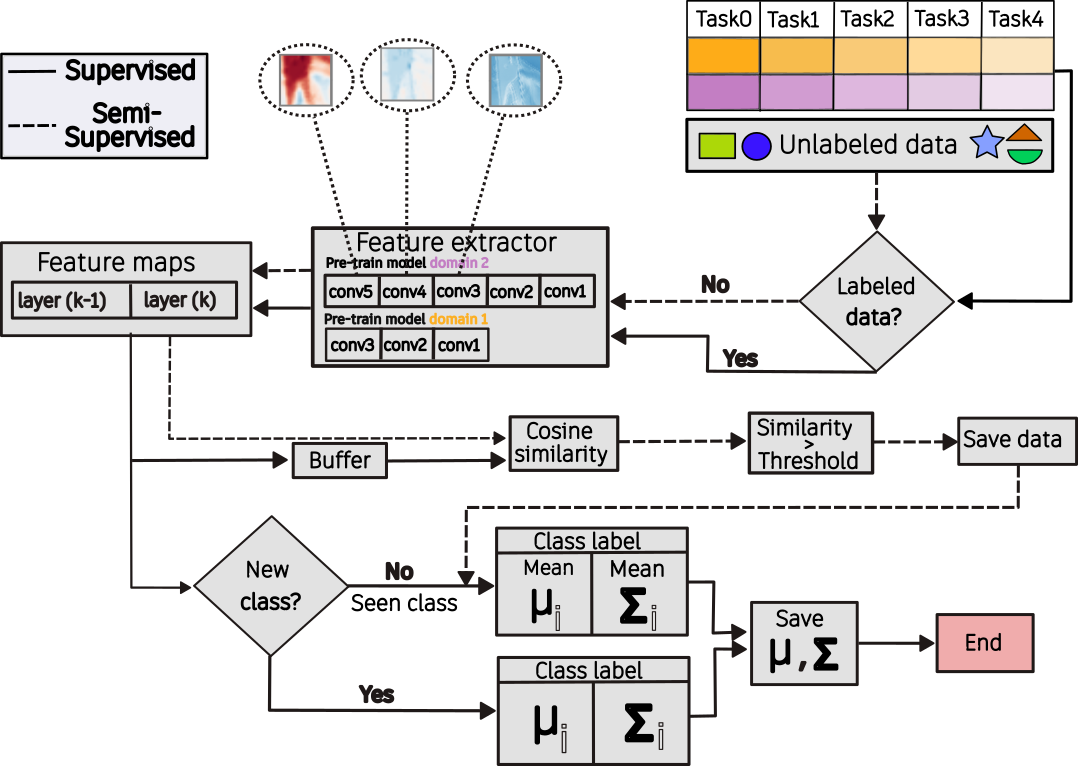}
	\caption{exFeCAM algorithm with intra-layer feature representation, online batchwise learning, and SSL capabilities for multimodal training. During training on multiple domains, the algorithm stores the mean and covariance matrices of the classes in a dictionary. In the testing phase, the algorithm uses these stored prototypes and employs the Mahalanobis distance to determine the prediction class. Supervised training is depicted with solid black lines, while unsupervised training is illustrated with dashed black lines.}
	\label{fig:fecam_algorithm}
\end{figure*}

\section{Methodology}
In this section, we discuss the modifications made to the FeCAM algorithm \cite{goswami2024fecam}. Originally, the algorithm models the feature distribution of each class using a multivariate normal distribution, $\mathcal{N}(\mu, \Sigma)$, and employs a Mahalanobis distance based metric for classifying the respective samples. Although the authors proposed several techniques (co-variance matrix approximation, shrinkage and normalization) to enhance and stabilize the training process, one significant limitation of FeCAM is it's requirement to store batches in temporary memory and conduct training only once all data for a particular task becomes available. This makes it unsuitable for online learning. To address this issue and improve the algorithm's capability for multimodal training, as well as to adapt to the semi-supervised nature of the dataset, we introduce several new features to the exFeCAM algorithm, which are detailed below.

\subsection{Online Batchwise Training}
Online training refers to incrementally updating a model in a batch-wise fashion, rather than processing the entire dataset for a given task at once. In this approach, each batch of data is passed through a feature extractor to obtain feature maps, which are then used to calculate the prototypes (mean and co-variance matrix) for each class in that batch. A dictionary is initialized with class labels as keys and their corresponding prototypes as values. There are two possible scenarios when updating this dictionary:
\begin{enumerate}
    \item If a class is encountered for the first time, a new entry is added with the class label as the key and the corresponding prototypes as the value.
    \item If the class has been seen before, the existing prototypes for that class are retrieved and updated by averaging the previous prototypes with the new ones from the current batch.
\end{enumerate}
This method avoids the need to store incoming data for each task, which is critical for CL objectives and is especially suitable for autonomous agents with limited cache memory. Learning prototypes in a batch-wise manner addresses the challenge of not having access to previous data points, making it a more practical solution for real-time applications.
\begin{table*}[]
    \centering
    \caption{Pre-train model architecture for both the domains of the custom multimodal dataset, domain 1 represents the image data and domain 2 signifies the sensor data. The linear layers for both the models are removed once the pre-training is completed . Here, "classes" refers to the total number of output classes during the pre-training phase.}
    \label{table_pretrainModel}
    \begin{tabular}{p{2.0cm}p{2.5cm}p{2.6cm}p{1.85cm}p{2cm}p{5cm}}
    \hline
    \textbf{Model type} & \textbf{Layer type} & \textbf{Activation function} & \textbf{Output channel/neurons} & \textbf{Batch Normalization} & \textbf{Other details}\\
    \hline\\
    & Conv2D &Relu &16 & Batchnorm2D &kernel:3, stride:1, padding:1 \\ 
    & Conv2D &Relu &32 & Batchnorm2D &kernel:3, stride:2, padding:1 \\
    & Conv2D &Relu &64 & Batchnorm2D &kernel:3, stride:2, padding:1 \\
    & Conv2D &Relu &64 & Batchnorm2D &kernel:3, stride:3, padding:0 \\
    Domain 1 & Conv2D &Relu &128 & Batchnorm2D &kernel:2, stride:2, padding:0 \\
    & Flatten &- &- & - &- \\
    & Linear &Relu & 2000& - & -\\
    & Linear &Relu & 1000& - & - \\
    & Linear &Relu & classes & - &-\\
    \hline\\
    & Conv1D &Leakyrelu (0.01) &512 & Batchnorm1D &kernel:3, stride:1, padding:1 \\
    & Conv1D &Leakyrelu (0.01) &256 & Batchnorm1D &kernel:3, stride:2, padding:1 \\
    & Conv1D &Leakyrelu (0.01) &128 & Batchnorm1D &kernel:3, stride:2, padding:1 \\
    Domain 2& Flatten &- &- & - &- \\
    & Linear &Relu & 256& - &-\\
    & Linear &- & classes& - &-\\
    \hline
    \end{tabular}
\end{table*}

\begin{algorithm*}[t]
\caption{exFeCAM Algorithm}
\label{alg:exfecam}
\textbf{Input:} Data stream $\left(D_0^0, D_0^1, \dots, D_1^0, D_1^1, \dots, D^{0:t}_n\right)$ \\
\textbf{Models:} Feature extractor $f_e(x; \theta_{e})$, exFeCAM model $\phi(x; \mu, \Sigma)$ \\
\textbf{Variables:} $k$: number of layers in feature extractor, $t$: total tasks, $n$: number of domains, $T$: total classes, $b$: batch size, $l, j$: iterative variables \\
\hspace*{2em} $(x_i, y_i) \in D_\text{train}^\text{Supervised}$, $(x_i^{'}) \in D_\text{train}^\text{Unsupervised}$, $(x_i^{''}, y_i^{''}) \in D_\text{test}$ \\
\hspace*{2em} $\mu, \Sigma \gets \{\}$

\textbf{Training Phase:}
\begin{algorithmic}[1]
\FOR{each experience $E_i \in \{E_0, E_1, \dots, E_t\}$}
    \STATE Temporary Buffer $(\epsilon)$ $ \gets \{\}$ \hfill \COMMENT{Temporary memory buffer for feature matching}
    \FOR{each class $y_i \in \{y_0, y_1, \dots, y_T\}$}
        \FOR{$j = 0$ to $b$}
            \STATE $\psi_{ij} \gets \large[ {f_e}{(x_i)}_{k-(l-1)}, {f_e}{(x_i)}_{k-(l-2)}, \dots, f_e(x_i)_k$ \large] \hfill \COMMENT{Concatenate intra-layer features}
            \STATE $\mu[i] \gets \frac{1}{b}\sum \psi_{ij}$ \hfill \COMMENT{Mean matrix}
            \STATE $\Sigma[i] \gets \text{Covariance}(\psi_{ij})$ \hfill \COMMENT{Covariance matrix}
            \STATE $\Sigma[i]_{S} \gets \text{Shrinkage}(\Sigma[i])$
            \STATE $\Sigma[i]_{N} \gets \text{Normalization}(\Sigma[i])_S$
            \IF {$\epsilon$ not empty}
                \STATE similarity $\gets$ cosine\_similarity$(\text{features} \in \epsilon, x_i^{'})$\hfill\COMMENT{Compute cosine similarity}
                \IF{similarity $\geq$ threshold}
                    \STATE Update $\mu[i]_S, \Sigma[i]_N$\hfill\COMMENT{Update the mean and cov matrix for each class}
                \ENDIF
            \ENDIF
        \ENDFOR
    \ENDFOR
\ENDFOR
\end{algorithmic}

\textbf{Test Phase:}
\begin{algorithmic}[1]
\FOR{each $x_i^{''} \in D_\text{test}$}
    \STATE $\phi(f_e(x_i^{''}); \mu, \Sigma) \gets (f_e(x) - \mu_i)^{T}(\sum_i)_N^{-1}(f_e(x) - \mu_i)$ \hfill\COMMENT{Calculate the Mahalanobis distance}
    \STATE $y_{\text{pred}} \gets \arg\min \phi(f_e(x_i^{''}); \mu, \Sigma), \forall y_i \in \{0, 1, \dots, T\}$\hfill\COMMENT{Assign the class with minimum Mahalanobis distance}
\ENDFOR
\end{algorithmic}
\end{algorithm*}

\subsection{Intra Layer Feature Representation}
Intra-layer feature representation (ILFR) leverages the power of intermediate layers that encodes the low and mid-level feature information present in the data extracted using a pre-trained model. Instead of relying solely on the final output layer of the feature extractor, the presence of multiple activation maps generalizes the learned knowledge and further improves the performance of the algorithm.  We argue that enriching the last layer representations with hierarchical intra-layer features increases robustness to domain shifts and thus improves generalizability to downstream continual tasks. These extracted features can be the class embeddings of a transformer encoder \cite{vaswani2017attention} or the flattened feature maps of a ResNet encoder \cite{he2016deep}.

There are two intuitive ways to promote intra-layer feature representations into the model: averaging the representations from the last "k" layers of the network or concatenating the representations from the last "k" layers. The first approach requires the output dimensions of all layers to be identical to facilitate summation. In our use case, we opt for concatenation, as the outputs from different layers have varying dimensions. Concatenation also preserves the rich, multi-scale features, unlike summation, which may result in information loss due to the merging of complementary features captured by different layers, potentially reducing the richness of the learned representations.

For the multimodal experiment, we pre-train the domain 1 feature extractor on gripper data of objects not present in the training dataset, while the domain 2 feature extractor is pre-trained on the Core50 dataset \cite{lomonaco2017core50}. For the VGGSound experiment, we use the ResNet18 architecture as the feature extractor, pre-trained on the ImageNet1k dataset \cite{russakovsky2015imagenet}. More details about the pre-trained model architecture are provided in Table \ref{table_pretrainModel}. 
\begin{figure*}
	\centering
	\includegraphics[width=0.9\textwidth]{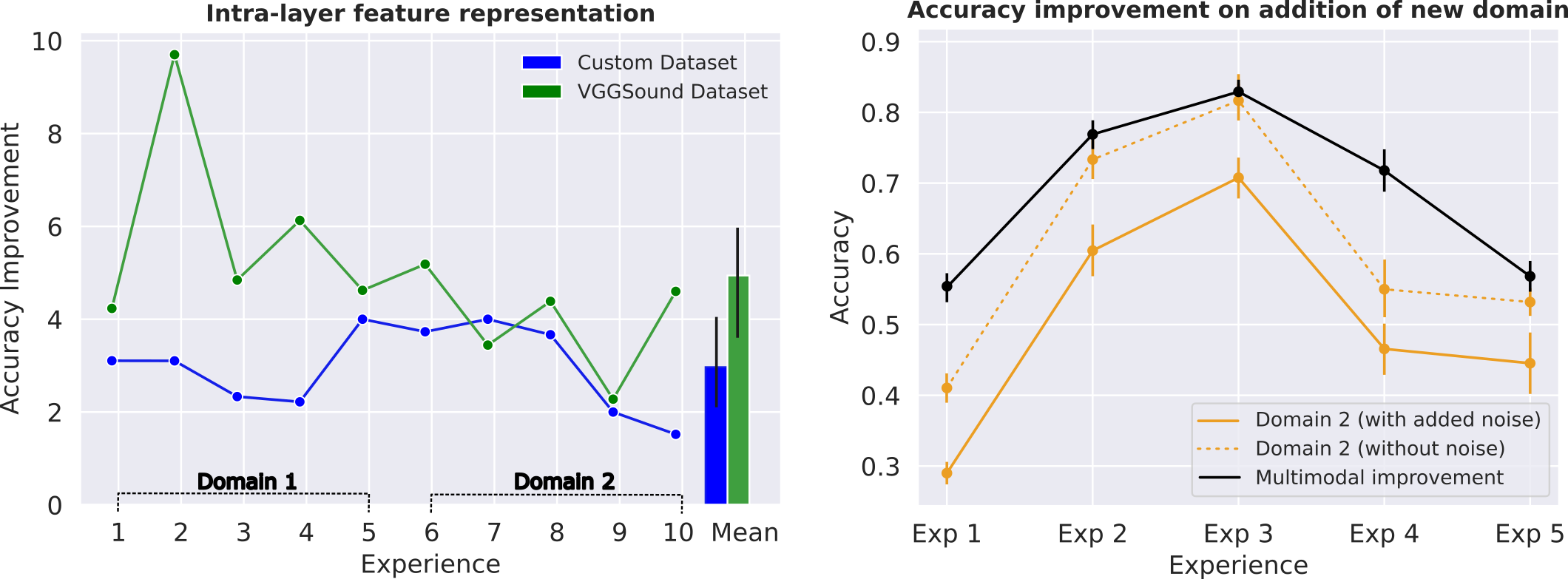}
	\caption{On the left: intra layer feature representation for the custom dataset and the VGGSound dataset, we compare the improvement in the accuracy of the FeCAM algorithm on addition of knowledge from the last as well the penultimate layers of the pre-trained feature extractors. On the right: the plot shows sensor data accuracy (in orange) with and without the addition of random gaussian noise, the solid line in black shows the improvement in the prediction accuracy for the different experiences after addition of the image domain.}
	\label{fig:ilfr}
\end{figure*}

\subsection{Semi-Supervised Learning}
We incorporate an SSL approach to enable our algorithm to leverage unlabeled data, thus enhancing it's overall generalization and improving performance on both previously encountered and newly introduced objects. To integrate SSL capabilities into the exFeCAM algorithm, we adopt a pseudo-labeling strategy, which allows us to organize the unlabeled data by assigning them dummy labels for further training. The pseudo-labeling process begins by randomly storing few feature maps of each class for every task in a temporary memory buffer. We then compute the cosine similarity between these reference feature maps and the feature maps generated from incoming unlabeled data, using the formula defined in Equation \ref{similarity}.

\begin{equation}\label{similarity}
similarity = \frac{x_1.x_2}{||x_1||_2||x_2||_2}
\end{equation}

Here, $x_1$ represents a reference feature map from the temporary memory buffer, while $x_2$ corresponds to the feature map extracted from the new unlabeled data. The notation $||.||_2$ indicates the L2 norm of the respective vectors. If the cosine similarity between the two feature maps exceeds a predefined threshold, the exFeCAM algorithm proceeds to train on the new feature map and it's assigned pseudo-label. These pseudo-labels are derived from the class labels associated with the most similar reference feature used during feature matching. The corresponding prototype values for that class, specifically the mean and covariance matrices are then updated according to the averaging rule discussed in the previous section.

To select an appropriate threshold for pseudo-labeling, we utilize the Optuna framework \cite{akiba2019optuna}, which employs bayesian optimization to efficiently explore the hyperparameter space and identify an optimal value. Once the pseudo-labeling process for a particular task is complete, the temporary memory buffer is discarded.

% till here

\begin{table*}[]
\caption{Comparison of the accuracy of the exFeCAM algorithm and the benchmark methods across two different multimodal datasets. The joint method represents a scenario where the algorithm has access to the entire dataset at once.}
\label{vgg_table}
\begin{center}
\begin{tabular}{|c||c||c||c||c||c||c|}
\hline
Dataset & Joint & Naive-add & Naive-concat & FeTrIL &Vanilla FeCAM & ExFeCAM\\[1mm] 
\hline
Custom multimodal & 94.65$_{\pm 1.29}$& 43.43$_{\pm 1.59}$ & 45.16$_{\pm 1.68}$ & 62.10$_{\pm 1.86}$ & 65.30$_{\pm 0.98}$ & 70.50$_{\pm 1.79}$ \\[1mm]
\hline
VGGSound & 92.48$_{\pm 2.31}$ & 29.78$_{\pm 1.76}$ & 30.94$_{\pm 1.81}$ & 38.21$_{\pm 0.89}$ & 39.96$_{\pm 1.16}$ & 45.96$_{\pm 1.98}$\\[1mm]

\hline
\end{tabular}
\end{center}
\end{table*}
\subsection{Multimodal Training}
We extend the traditional class-incremental CL framework to a more practical and challenging semi-supervised continual learning (SSCL) scenario. In this setting, the data distributions are designed to mirror real-world correlations among object classes by incorporating both labeled and unlabeled samples. For our experiments, we use a data split consisting of 70\% labeled and 30\% unlabeled data, applied under two protocols (random and unseen) as described in the previous section. At each task \textit{n}, we represent the batches of labeled data as

$$
X_n = \{(x_b, y_b) : b_i \in (1, ..., b), \, y_b \in \tau_n\}
$$

and the batches of unlabeled data as

$$
X_n^{'} = \{x_b^{'} : b_i \in (1, ..., b)\},
$$

where $b$ denotes the batch size and $\tau_n$ indicates the set of classes associated with task $n$. The main goal during each task is to train a model that can accurately classify any query instance drawn from the union of all classes encountered in the current and preceding tasks, i.e., $\tau_0 \cup \tau_1 \cup \dots \cup \tau_n$. As training progresses, the model encounters new classes incrementally within each domain, as illustrated in Figure \ref{fig:multimodalData}, where each domain corresponds to a different data modality. For each task, input data are passed through a feature extractor that produces feature maps from various layers of the network. These feature maps are standardized to a fixed dimension and normalized to 0–1 range before being fed into subsequent stages of the pipeline. These processed feature representations are then used to incrementally train the exFeCAM algorithm. To support the semi-supervised aspect of learning, a small subset of feature maps corresponding to the objects in each task is stored temporarily in a memory buffer. This buffer is utilized for pseudo-labeling of unlabeled data, as discussed in earlier sections.

During the training phase, the algorithm computes the mean and covariance matrix for each batch of data and apply various techniques such as covariance matrix approximation, shrinkage, and normalization to ensure robust and stable learning, as described in the paper \cite{goswami2024fecam}. In the testing phase, feature maps for the test instances are extracted using the appropriate feature extractor, and classification is performed by computing the Mahalanobis distance between each test sample and the class prototypes learned during training. The class corresponding to the minimum distance is selected as the predicted label. Further details about the algorithm implementation are provided in the corresponding pseudo-code.

\begin{figure*}
	\centering
	\includegraphics[width=\textwidth]{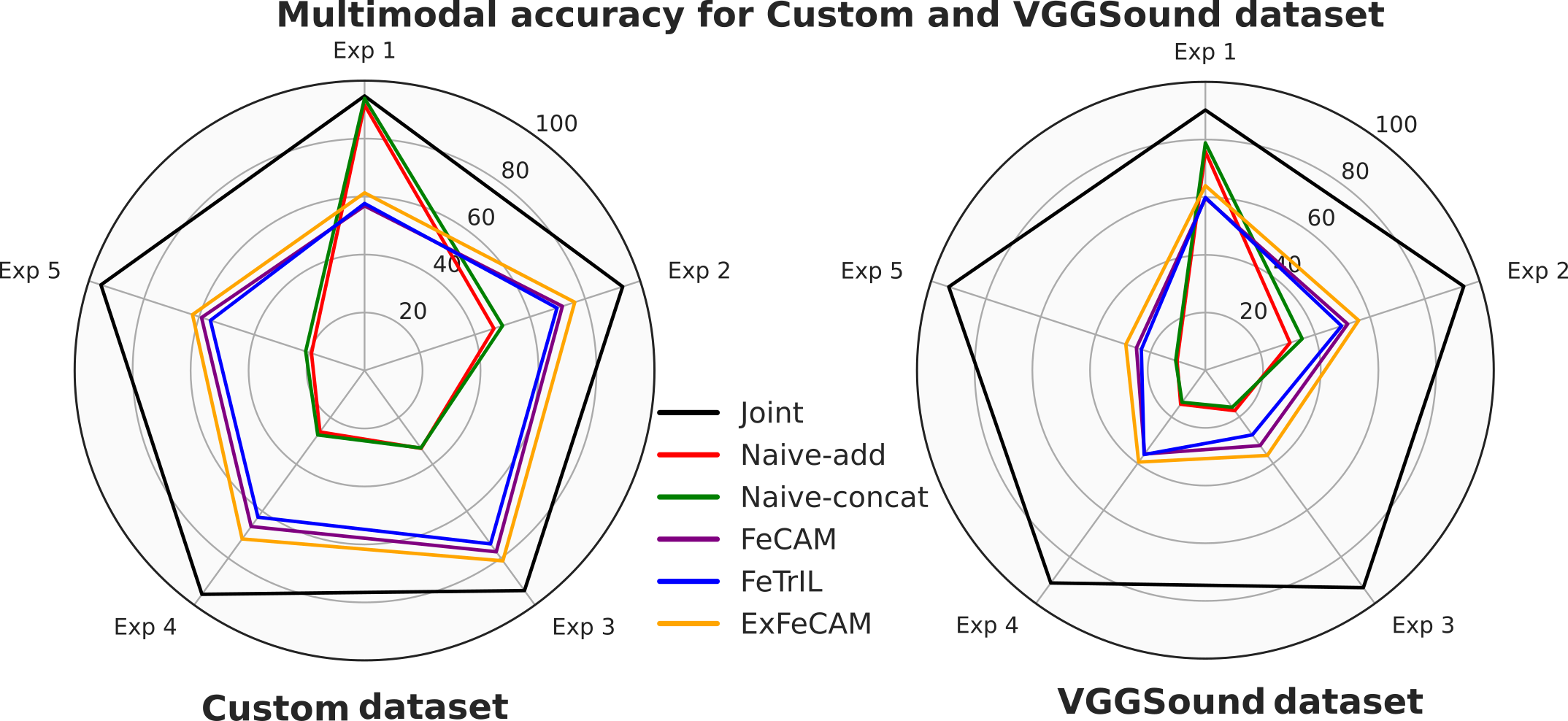}
	\caption{On the left, the radar plot illustrates the accuracy of various algorithms after incremental training on each experience for the custom multimodal dataset. The accuracy per experience represents the average performance across two domains: visual and sensor signals. On the right, the radar plot presents the per-experience accuracy for the VGGSound dataset, where the two domains consist of visual and audio data.}
	\label{fig:spider_plot}
\end{figure*}

\section{Results}
To evaluate the performance of our algorithm and the effectiveness of the proposed modifications, we assess the exFeCAM algorithm on both a custom multimodal dataset and the VGGSound dataset. It is important to note that the primary focus of this paper is not to establish a new level of accuracy for object classification, but rather to demonstrate the effectiveness of CL in a challenging multimodal training setting. All experiments are conducted three times, and we report the mean and standard deviation of the relevant metrics.

Table \ref{vgg_table} compares the accuracy of various algorithms across the two multimodal datasets. Our proposed method achieves a significant improvement over all benchmark algorithms, which can be attributed to the integration of SSL and intra-layer feature representation (ILFR) capabilities in our architecture. Additionally, the vanilla FeCAM algorithm produces competitive results, likely due to it's offline training mechanism. In contrast, our algorithm employs batch-wise online training, making it more relevant for real-world scenarios. FeTrIL algorithm also achieves a comparable accuracy, mostly attributed due to the presence of the pseudo-rehearsal technique to compensate forgetting of previous knowledge.

Figure \ref{fig:ilfr} (left) illustrates the enhancement in our algorithm's prediction accuracy with the integration of intra-layer feature representation (ILFR). The accuracy is measured after the model has been incrementally trained on all tasks. For the custom dataset (represented in blue), the first five experiences depict the percentage improvement in domain 1 data (vision), while the subsequent five experiences reflect the accuracy gains in domain 2 (tactile sensor). Each experience consists of two distinct classes for both domains. On average, incorporating ILFR results in an overall accuracy improvement of approximately 3.12 $\pm$ 1.09 compared to the baseline without ILFR. A similar trend is observed in the VGGSound dataset (also shown in blue), where the initial five experiences correspond to image data, followed by five experiences with the associated audio data. In this case, the overall accuracy improvement is approximately 4.90 $\pm$ 1.20, further validating the benefits of ILFR.

Figure \ref{fig:ilfr} (right) demonstrates the advantage of training on a new domain when the information from the original domain (signals from the tactile sensor) is noisy. To simulate this scenario, gaussian noise is added to the sensor signal data. The solid orange line indicates the accuracy across different experiences for the signal domain with noisy sensor data, while the dashed orange line shows the original accuracy without noise. The solid black line indicates the improvement in accuracy when image data is incrementally introduced to mitigate the performance degradation caused by noisy sensor signals.

The CL capabilities of exFeCAM, along with the benchmark algorithms, are presented in Figure \ref{fig:spider_plot}. The spider plot depicts the combined accuracy for both domains, demonstrating that our proposed algorithm consistently outperforms all other methods across all experiences. The naive-concat and naive-add benchmark methods perform well in the initial experience but suffer significant accuracy degradation as the number of classes increases due to catastrophic forgetting. The joint method, which trains on all data at once, serves as an upper bound on per-experience accuracy. The FeTrIL algorithm initially shows results comparable to vanilla FeCAM, but it's performance declines in later tasks, likely due to the moderate quality of the pseudo-rehearsal samples generated during training on new tasks.

\begin{figure}
	\centering
	\includegraphics[]{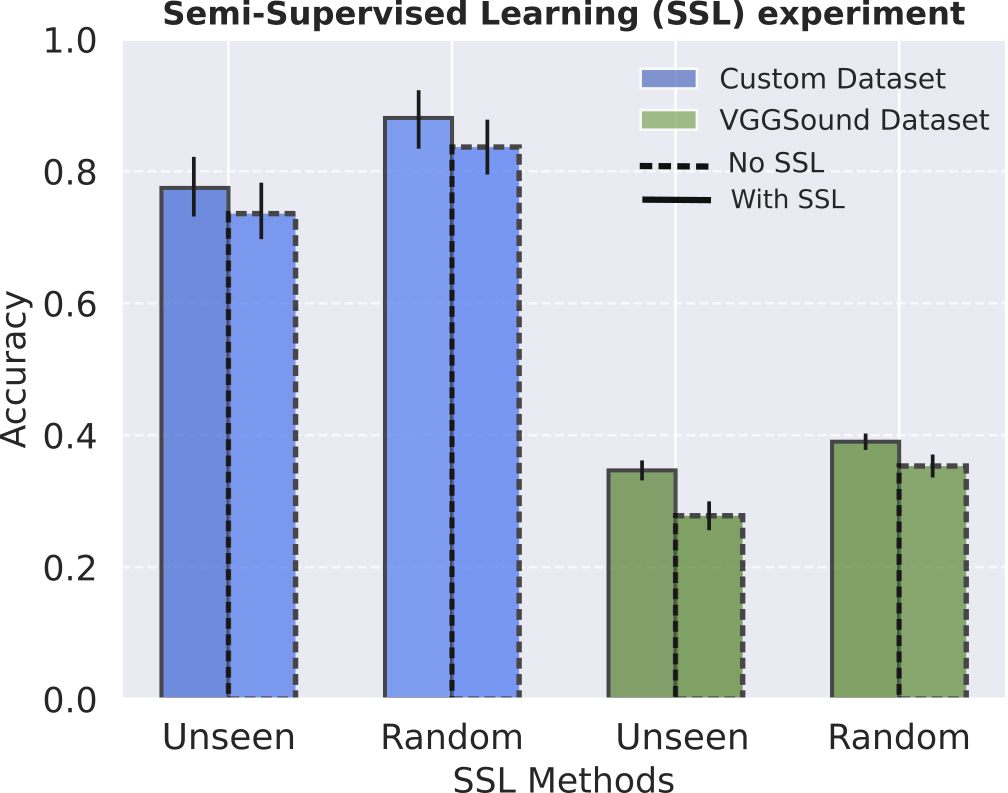}
	\caption{Accuracy of the exFeCAM algorithm across two different SSL scenarios. The blue plots correspond to the custom dataset, while the green plots represent the VGGSound dataset. Bar plots with solid lines indicate the condition where SSL is integrated into the algorithm, whereas dotted lines represent the condition without SSL.}
	\label{fig:ssl_all}
\end{figure}

Additionally, we conducted a separate SSL experiment exclusively on the visual modality from both datasets, implementing two distinct types of experiments within the SSL protocol, as described in Section \ref{eval_metric}. Figure \ref{fig:ssl_all} presents the accuracy results for the custom dataset and the VGGSound dataset under both scenarios. The blue bar plot with a solid boundary represents the accuracy of the exFeCAM algorithm for the \textit{unseen object} and \textit{random object} scenarios with SSL enabled, while the dotted-boundary plot indicates the accuracy for the same scenarios without SSL. Similarly, the accuracy results for the VGGSound dataset are depicted using green bar plots. However, the overall accuracy of the VGGSound dataset is lower compared to the custom dataset, possibly due to the increased complexity and diversity of backgrounds in the extracted frames, whereas the custom dataset primarily consists of images with a relatively uniform background.

\begin{figure*}
	\centering
	\includegraphics[width=\textwidth]{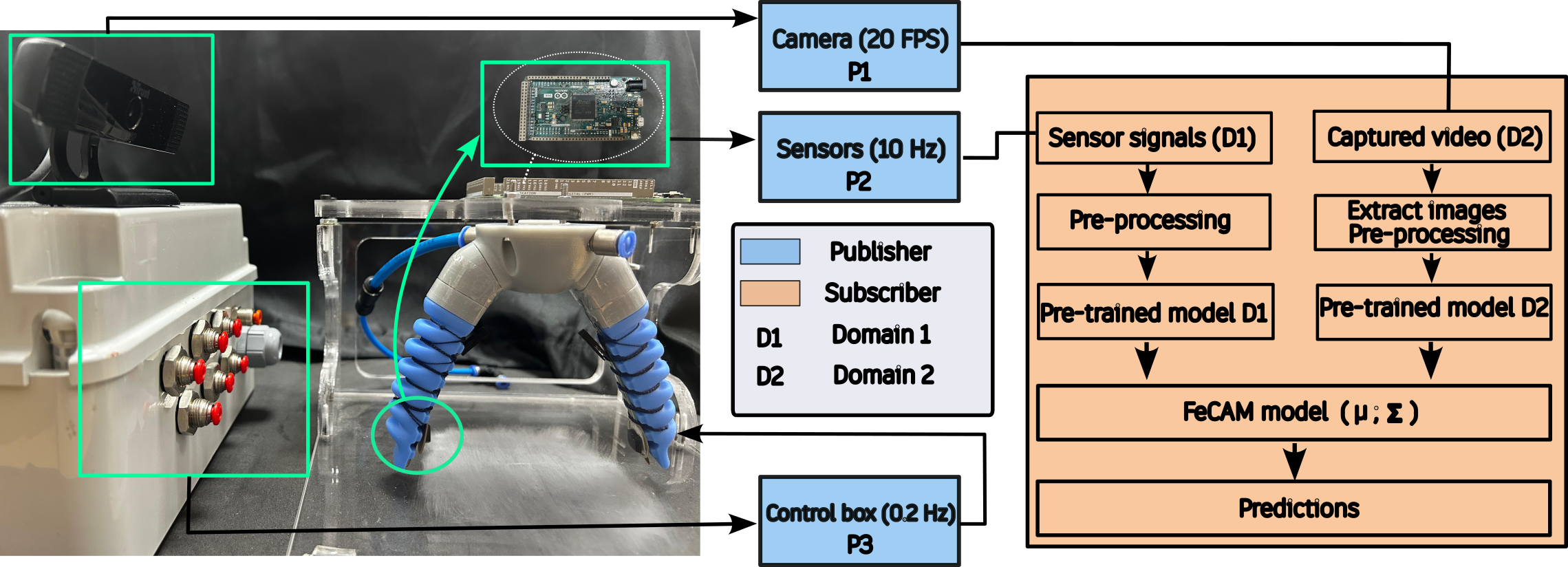}
	\caption{The real-time multimodal experiment using ROS involves three publishers and one subscriber node. The publishers (P1, P2, P3) are depicted in blue, while the subscriber is shown in yellow. P1 operates at a frequency of 20 Hz, P2 operates at 10 Hz, and P3 operates at 0.2 Hz. The output from P3 is used to actuate a soft finger equipped with sensors. The outputs from P1 and P2 are fed to the subscriber, where they undergo pre-processing, feature extraction using a pre-trained network, and finally to the FeCAM algorithm to make the predictions using the saved mean and co-variance matrices.}
	\label{fig:ros_flowchart}
\end{figure*}

\section{Real-Time Evaluation With ROS}\label{ros_application}
Robot Operating System (ROS) \cite{koubaa2017robot} is a comprehensive framework that comprises a suite of tools, libraries, and protocols aimed at facilitating the development of various robotic systems. The system manages the creation and control of communication between a robot’s peripheral modules, such as sensors, cameras, and actuators, thereby enabling the seamless integration of these components with parallization capabilties.
\begin{figure*}
	\centering
	\includegraphics[width=\textwidth]{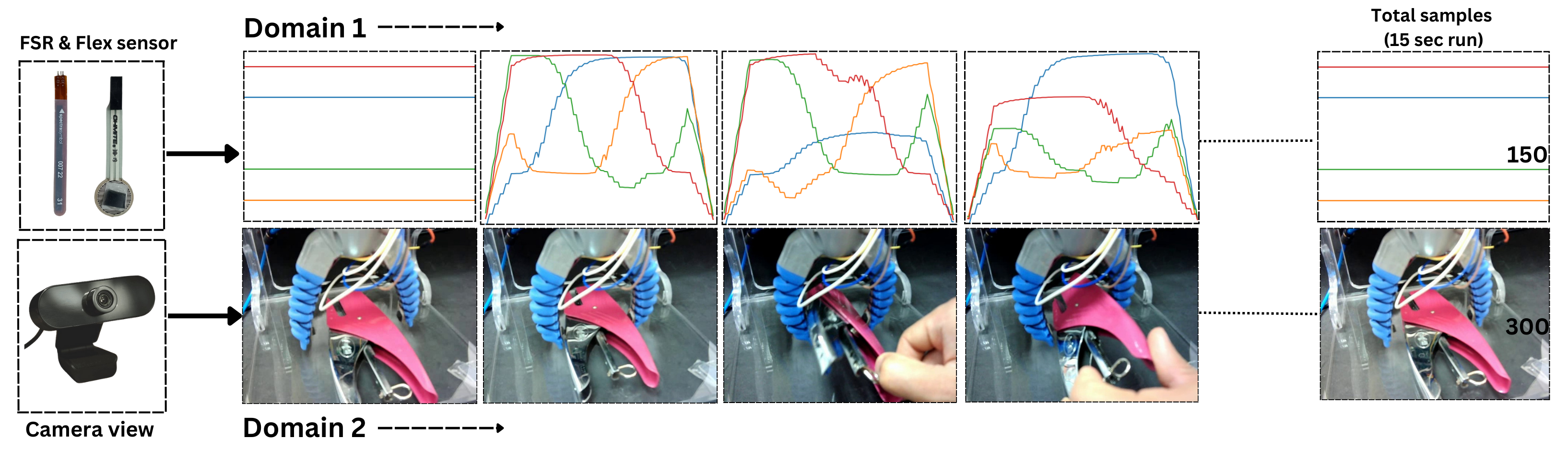}
	\caption{Real-time images of the object from the camera's perspective, along with the sensor data from four sensors attached to the soft finger, upon it's actuation. These images and sensor readings are directly fed into the FeCAM model after resizing them to the desired dimension. For a single 15-second run, there are a total of 150 sensor signals and 300 images, representing the number of raw data points and corresponding images of the object.}
	\label{fig:realtime_training}
\end{figure*}
\subsection{ROS Setup}

ROS is primarily designed to function optimally on Ubuntu, while only partially compatible with other operating systems. For our case we use ROS on Ubuntu 20.04 on a Intel Core i7 7th gen CPU $@$ 2.80 GHz equipped with Nvidia GTX 1080 GPU. We use the Noetic Ninjemys distribution of ROS 1 \cite{koubaa2017robot} in our experiment. A detailed explanation about the workflow of the project is provided in the flowchart in Figure \ref{fig:ros_flowchart}. 

\begin{figure}
	\centering
    \includegraphics[]{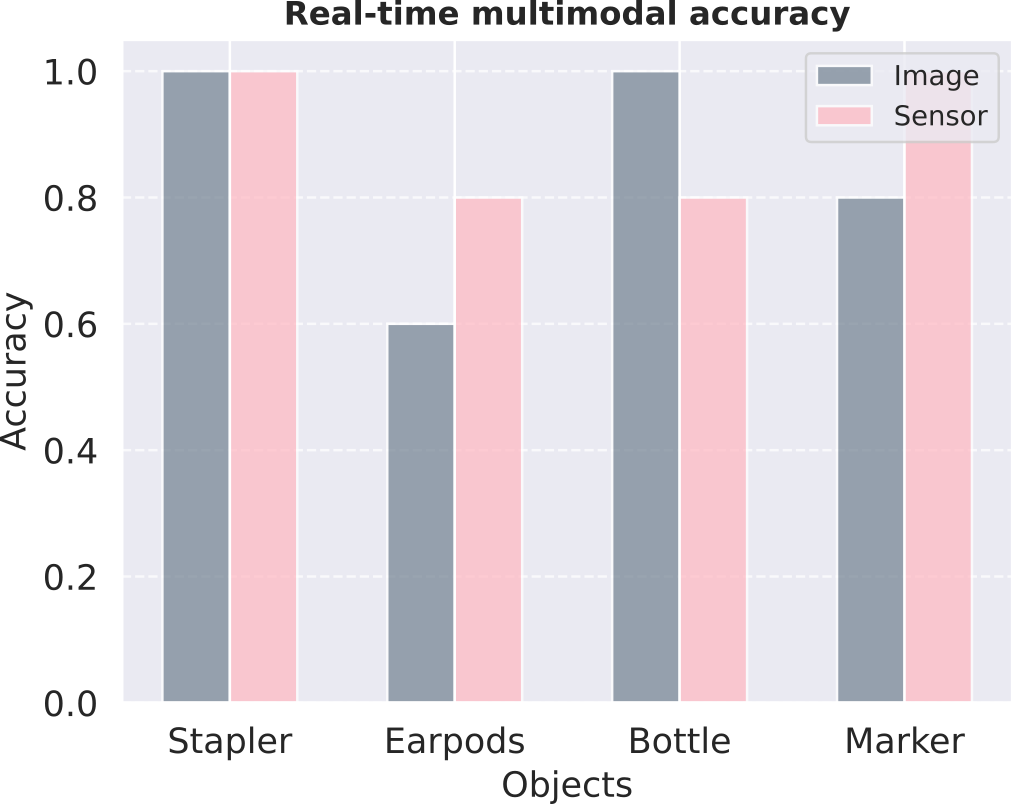}
	\caption{Real-time object classification accuracy of the two modalities for four different objects. The grey plot represents the prediction accuracy in the image domain, while the pink plot highlights the accuracy for the sensor domain.}
	\label{fig:real_exp}
\end{figure}
The first publisher (P1) publishes the sensor signals coming from the four sensors attached to the soft finger, the second publisher (P2) publishes the video coming from the camera and the third publisher (P3) publishes the command pressure to the control box which inturn actuates the finger. The control frequency of P1 and P2 are 10hz and 20hz, whereas the actuation frequency of P3 is 0.2hz. We further observed that actuating the finger at a more higher frequency constrained it from properly holding the object and thus effected the quality of the generated data from the sensors. The data from P1 and P2 goes to the subscriber which then pre-process the data, extracts the frames from the video in case of P2 and sends it to the respective feature extractors. The output of the feature extractor is a 384-dimensional feature map for both the domains which then goes as an input to the exFecam model for further processing and prediction.

\subsection{Real-Time Experiment}
To assess the performance of our algorithm, we conduct a quasi real-time experiment using a soft pneumatic gripper equipped with two flex sensors and two force sensors. During the evaluation phase, data from the tactile sensors and the camera are collected in real-time for approximately 15 seconds. These multimodal data are treated as two distinct domains and are sent through the dedicated pipeline in parallel for further processing and subsequently to the exFecam model for predictions as shown in Figure \ref{fig:ros_flowchart}. To generate the predictions we utilize the stored mean and the co-variance matrix computed during the training phase. Impressively, the mean and covariance matrix occupy only 22.5 MB of memory, making it quite efficient compared to other CL strategies.

However, since the model was initially trained on images of the objects in a different frame of reference, where the operator extends his arm and smoothly rotates the object in front of the camera. In contrast, during the real-time experiment, the objects are positioned near the soft finger, and a camera captures images from a top-down view. This difference sometimes leads to incorrect predictions by the model. To address this issue, for some objects, we occasionally need to refine the model's knowledge by conducting a brief training session (lasting around 82 seconds) specifically for that object. Figure \ref{fig:realtime_training} shows a snapshot of the real-time data collection from both the tactile sensors and the camera's perspective. The image shows the progression of the sensor signal and the camera image upon actuating the finger. For each cycle the sensors collect 150 raw values, which are transformed into a single 600-dimensional vector, whereas the camera collects around 300 images of the object from various orientations within the 15-seconds interval.

As discussed earlier, a significant advantage of domain-incremental CL training is it's ability to incorporate new data modalities incrementally. This is particularly beneficial in scenarios where the input from one sensor becomes unreliable, either due to obstructions in the environment that limit it's field of view or due to noise introduced by external factors. This advantage is further illustrated in Figure \ref{fig:real_exp}, where we evaluate object classification accuracy in real-time for four different objects: stapler, earpod, bottle, and marker pen. The results reveal that for the "earpod" and "marker pen", the tactile sensor modality achieves notably higher classification accuracy than the corresponding visual modality. This discrepancy is likely caused by occlusions from the operator’s hand during image capture or limitations in the visual feature extractor’s ability to accurately represent small objects. In such cases, the availability of a second modality (here tactile sensor) effectively compensates for the missing or unreliable information, thereby improving overall classification performance. All the experiments were repeated 5 times for each of the four objects in real-time, and the final prediction accuracy was computed accordingly.

\subsection{Real World Application}
A real world application for such a technique can be in companion robots \cite{broadbent2023enhancing}, where the robot learns new tasks incrementally by interacting with it's environment and recognizing different objects using information from multiple sensors available to it. The use of multiple sensors is particularly advantageous in scenarios where a single sensor cannot provide complete information about an object. For instance, a vision camera might struggle to differentiate between two objects with similar structures but with different textures or materials, with one being much softer than the other. In such cases, information from a tactile sensor can be extremely beneficial. Our proposed framework simplifies the addition of new modalities by treating each as a new domain, eliminating the need to re-train the entire algorithm from scratch each time.

\section{Conclusion and Discussion}
In this paper, we present a method for incrementally learning new data modalities without the need to re-train the model from scratch each time a new modality becomes available. Our approach combines class-incremental CL to progressively learn new object classes and domain-incremental CL to incorporate additional modalities for existing classes. To support this dual-incremental setup, we build upon the original FeCAM algorithm by introducing key enhancements, including online batch-wise training, intra-layer feature representation, and SSL capabilities, resulting in a more robust and versatile framework.

Our method employs cosine similarity based feature matching to generate pseudo-labels for unlabeled data, enabling the model to effectively leverage both labeled and unlabeled examples during training. We evaluate the proposed algorithm on two multimodal datasets of varying complexity and demonstrate it's ability to incrementally learn new classes in different domains. To better understand the contribution of individual components, we also conduct a comprehensive ablation study. To validate real-world applicability, we perform a real-time experiment using a soft pneumatic gripper equipped with four tactile sensors and an external camera. This experiment highlights the practical advantages of our multimodal CL approach in a realistic setting.

As part of our future work, we plan to further improve the exFeCAM algorithm by incorporating forward and backward knowledge transfer, either through interactions between class prototypes or by training a multilayer perceptron (MLP) network to model these interactions. Additionally, instead of relying on a simple average of predictions from both modalities, we aim to adopt a more refined approach by using a weighted averaging scheme, where a separate network will estimate the confidence of each prediction to determine the appropriate weights.

\medskip

% Acknowledgements
\medskip
\section*{Acknowledgements} 
We acknowledge the contribution from the Italian National Recovery and Resilience Plan (NRRP), M4C2, funded by the European Union–NextGenerationEU (Project IR0000011, CUP B51E22000150006, "EBRAINS-Italy")
\section*{Conflicts of Interest}
The authors declare that they have no known competing financial interests or personal relationships that could have appeared to influence the work reported in this paper.
\section*{Data Availability Statement}
Data will be made available from the corresponding author upon reasonable request.
\section*{Ethical Approval Statement}
All experiments were conducted solely by the author, and no other human participants were involved in the study. The author therefore gives informed written consent and confirms that no ethics committee approval is required, as per university guidelines.

% References
\medskip

% Use the following code if you wish to generate your bibliography with BibTeX;
% replace the string "MSP-template" below with the name(s) of
% the BibTeX data base(s) you want to use.
% The resulting bibliography-output (the content of the .bbl file)
% must be pasted back into this file before submission.
% Please also include your BibTeX data base file(s) in your submission
% so that we can re-run BibTeX if necessary.
%
\bibliographystyle{MSP}
\bibliography{multimodal_ref}

\end{document}